\documentclass{article}
\usepackage{ijcai24}

\usepackage{times}
\usepackage{soul}
\usepackage{url}
\usepackage[utf8]{inputenc}
\usepackage[small]{caption}
\usepackage{graphicx}
\usepackage{amsmath}
\usepackage{amsthm}
\usepackage{booktabs}
\usepackage{algorithm}
\usepackage{algorithmic}
\usepackage[switch]{lineno}

\usepackage{amsfonts}
\usepackage{amsmath}
\usepackage{booktabs}
\usepackage{multirow}
\usepackage[colorlinks,linkcolor=red,citecolor=black]{hyperref}
\usepackage[capitalize]{cleveref}

\title{DiffSpeaker: Speech-Driven 3D Facial Animation with Diffusion Transformer}

\author{
Zhiyuan Ma$^{1,2}$
\and
Xiangyu Zhu$^3$\and
Guojun Qi$^{4}$\and
Chen Qian$^5$\and \\
Zhaoxiang Zhang$^{1,2,3}$\and
Zhen Lei$^{1,2,3}$\thanks{Corresponding author.}\\
\affiliations
$^1$The Hong Kong Polytechnic University\\
$^2$Center for Artificial Intelligence and Robotics, HKISI CAS\\
$^3$State Key Laboratory of Multimodal Artificial Intelligence Systems, CASIA\\
$^4$OPPO Research 
$^5$SenseTime Research\\
\emails
zm2354.ma@connect.polyu.hk,
xiangyu.zhu@nlpr.ia.ac.cn,
guojunq@gmail.com, \\
qianchen@sensetime.com,
\{zlei, zhaoxiang.zhang\}@nlpr.ia.ac.cn
}

\begin{document}

\maketitle

\begin{abstract}

Speech-driven 3D facial animation is important for many multimedia applications. Recent work has shown promise in using either Diffusion models or Transformer architectures for this task. However, their mere aggregation does not lead to improved performance. We suspect this is due to a shortage of paired audio-4D data, which is crucial for the Transformer to effectively perform as a denoiser within the Diffusion framework. To tackle this issue, we present DiffSpeaker, a Transformer-based network equipped with novel biased conditional attention modules. These modules serve as substitutes for the traditional self/cross-attention in standard Transformers, incorporating thoughtfully designed biases that steer the attention mechanisms to concentrate on both the relevant task-specific and diffusion-related conditions.  We also explore the trade-off between accurate lip synchronization and non-verbal facial expressions within the Diffusion paradigm. Experiments show our model not only achieves state-of-the-art performance on existing benchmarks, but also fast inference speed owing to its ability to generate facial motions in parallel. Our code is avalable at \href{https://github.com/theEricMa/DiffSpeaker}{https://github.com/theEricMa/DiffSpeaker}.


\end{abstract}

\begin{figure}[t]
\centering
\includegraphics[width=1\linewidth]{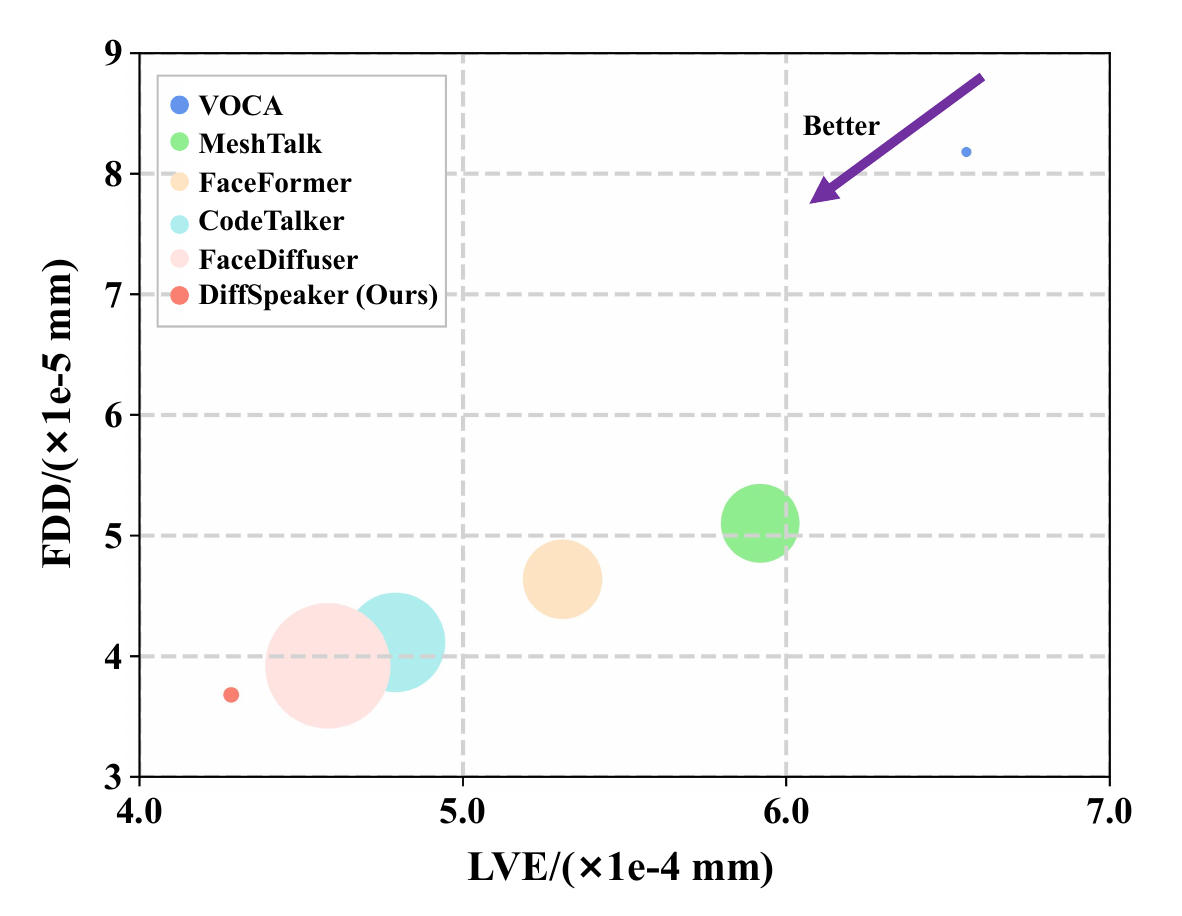} 
\caption{
    DiffSpeaker outperforms in quality (FDD, LVE metrics in Table~\ref{tab:quantitative_comparsion}) and speed, 
    with circle size indicating average inference latency for 10-90s duration of audio clips. 
    Despite Diffusion-based generation, it ensures \textbf{fast inference} (details in Figure \ref{Fig:latency}).
}
\label{Fig:teaser}
\end{figure}

\section{Introduction}
 Speech-driven 3D facial animation is a crucial component in many applications, such as in virtual assistants, video games, and movies productions. The goal is to create facial movements that are consistent and realistic with the driving speech. In other words, the facial animations should accurately mirror the tone and rhythm of the speech, giving the digital human a natural and believable speaking dynamics, so that users can interact with it in a more comfortable and natural way. To achieve this, researchers have been working on developing sophisticated algorithms and techniques that can accurately generate facial movements in sync with speech inputs.


In the realm of deep learning, there have been considerable progresses in synthesizing facial movements directly from audio in an end-to-end fashion. Initial methods were constrained by the use of a basic sliding-window approach for audio input, which often led to a narrow range of generated facial motions \cite{cudeiro2019capture,richard2021meshtalk}. Recent progress in deep learning has embraced the impact of the Transformer architecture, as originally introduced by \cite{vaswani2017attention}. The FaceFormer \cite{fan2022faceformer} exemplifies this trend, applying the extensive contextual capabilities of Transformer for auto-regressive facial motion generation. However, while the Transformer model has been successful in speech-driven 3D facial animation~\cite{xing2023codetalker,peng2023selftalk}, its traditional application using deterministic regression may not be the best approach. Deterministic regression assumes a fixed mapping between input speech and facial movements, but human speech and facial expressions are variable and dynamic, making it difficult to accurately capture their relationships with a single mapping. To overcome this limitation, a novel approach that can effectively capture the complex relationships between speech and facial movements is needed. Conditional probabilistic models, which learn a probabilistic mapping between speech and facial movements, provide a promising approach to speech-driven 3D facial animation.

In recent years, Diffusion models~\cite{ho2020denoising} have made considerable strides in generating various types of media, including images~\cite{rombach2022high} and videos~\cite{luo2023videofusion}. Diffusion models differ from traditional probabilistic methods, such as VAEs~\cite{kingma2013auto} and GANs~\cite{goodfellow2020generative}, in that they gradually remove noise from a signal rather than learning a mapping directly from a noise distribution to the data distribution. This approach allows Diffusion models to capture the entire distribution of the data, rather than just the modes, resulting in a higher degree of diversity in the generated output~\cite{dhariwal2021diffusion}. Diffusion models can also take advantage of additional inputs, such as class label~\cite{ho2022classifier} or text prompts~\cite{rombach2022high} to guide the generation process, producing data that meet the conditions provided. Given the performance of Diffusion models in generating various outputs based on input conditions, it seems beneficial to formulate our task as a conditional Diffusion model. There is already a body of work \cite{stan2023facediffuser,stan2023facediffuser,thambiraja20233diface} that supports the effectiveness of this probabilistic approach in capturing the nuanced variations of facial expressions that accompany speech. Although there have been improvements, current techniques still tend to create facial animations in brief segments, relying heavily on the sequential processing capabilities of GRUs \cite{cho2014learning} or convolutional networks. These architectures struggle with handling extensive context, unlike Transformers, which excel in this regard.

However, incorporating the Transformer architecture within a Diffusion framework poses a significant challenge, as the model is expected to denoise facial motion sequences over their entire length, which is a demanding task for the data-intensive attention mechanism. To address this, we develop a specialized method that incorporates biased conditional self/cross-attention. This design takes into account both the diffusion steps and speaking style, integrating them as conditioning tokens in self/cross-attention process, while also applying fixed biases for steering the attention within the task-specific requirement. This dual approach efficiently reduces the data intensity of self/cross-attention in the Diffusion-based Transformer architecture and proves effective for our task, which involves limited audio-4D data of brief duration. Moreover, our research explores the trade-off between precise lip synchronization and the creation of facial expressions not strictly linked to spoken audio. The successful fusion of Diffusion-based generation with the Transformer architecture not only yields superior performance on existing datasets but also facilitates the concurrent processing of entire audio clips, resulting in faster generation speeds compared to traditional sequential methods, as shown in Figure~\ref{Fig:teaser}.

In summary, our work presents a novel approach to integrating the Transformer architecture with a Diffusion-based framework, specifically tailored for speech-driven 3D facial animation. Our key contribution is a unique biased conditional self/cross-attention mechanism, which effectively addresses the challenges posed by training the Diffusion-based Transformer with limited and short audio-4D data. By combining Transformer with Diffusion-based generation, we achieve faster generation speeds, surpassing existing methods in performance and efficiency.

\section{Related Work}

\subsection{Speech-Driven 3D Facial Animation with Deterministic Mapping}

Speech-driven 3D facial animation seeks to generate lifelike facial movements from audio speech inputs, aiming for a synchronization that captures the tone, rhythm, and dynamics of the speech. Over the years, research has focused on algorithms for this purpose, most of which follow the paradigm of deterministic mapping, i.e. one audio corresponds to one facial motion. Early efforts involved crafting artificial rules that link speech sounds (phonemes) to facial movements (visemes), with systems measuring the influence of phonemes over visemes \cite{massaro201212,edwards2016jali}. The limited performance of rule-based methods have led to a shift towards machine learning techniques. Taylor et al. \cite{taylor2017deep} employed deep neural networks to transform phoneme transcriptions into facial animation parameters, while Karras et al. \cite{karras2017audio} used convolutional neural networks to animate faces directly from audio data. Recent studies relevant to our work have trained neural networks on audio datasets and 3D facial meshes from various individuals, aiming for high-resolution, vertex-level animations. VOCA \cite{cudeiro2019capture} inputs raw audio and speaker style, represented by subject identity, and utilizes temporal convolutions to animate a static mesh template. MeshTalk \cite{richard2021meshtalk} creates a categorical latent space to distinguish between audio-correlated and uncorrelated facial movements. They tend to overlook long-term audio context due to their reliance on short audio windows. FaceFormer~\cite{fan2022faceformer} addresses this by implementing the Transformer architecture \cite{vaswani2017attention}, which encompasses a broader context range and enables comprehensive facial animation. To tame the data-intensive attention operation~\cite{dosovitskiy2020image} for this task with limited audio-4D paired data, they proposed the static attention bias adopted from AliBi~\cite{press2021train}. Other studies~\cite{peng2023selftalk} have incorporated multi-modal losses with speech-to-text models \cite{baevski2020wav2vec}. Recent studies have recognized the inherent one-to-many relationship in the task, where a single speech input can correspond to  multiple facial motions . CodeTalker \cite{xing2023codetalker} successfully learned this complex data distribution using a quantized codebook, leading to noticeable performance improvement. This progress confirms that exploring one-to-many relationship is a promising avenue.

\subsection{Probabilistic Mapping with Diffusion Models}

The one-to-many nature has been extensively investigated in other tasks, using generative models that draw samples from a conditional probability distribution. Diffusion models \cite{ho2020denoising,song2020denoising}, which belong to the same family of generative approaches as GANs \cite{goodfellow2020generative} and VAEs \cite{kingma2013auto}, use a controlled Markov process to transform a Gaussian distribution into the target conditional data distribution. This is done by incorporating specific conditions into the model architecture, guiding the generative process. Such a strategy has allowed Diffusion models to capture complex data distributions with impressive effectiveness, as seen in applications like categorical generation \cite{ho2022classifier}, text-conditioned generation~\cite{rombach2022high}, or audio-conditioned generation~\cite{alexanderson2023listen,qi2023diffdance}. Within the scope of speech-driven 3D animation, researchers have widely explored its possibility to compete with existing methods.  In the field of speech-driven 3D facial animation, the research community has actively investigated how it performes against traditional techniques. FaceDiffuser \cite{stan2023facediffuser} represents an initial foray into this area, employing a Diffusion-based generative framework alongside GRU to individually address audio segments. Additionally, there have been applications of Diffusion models for the concurrent generation of head poses \cite{park2023df,sun2023diffposetalk}, customization to individual users \cite{thambiraja20233diface}, and methods like Diffusion distillation \cite{chen2023diffusiontalker} to accelerate the generative process. Some concurrent studies ~\cite{park2023said,aneja2023facetalk,zhao2024media2face} focus on blendshape-level animation with customized datasets. There is a growing need to explore how a Diffusion-based generative framework performs against standard benchmarks in our task, ensuring unbiased comparisons.

\section{Methods}

Speech-driven 3D facial animation aims to  generate facial motion $\mathbf{x}^{1:T}$  given the audio $\mathbf{a}^{1:T}$ of time duration $T$ and the speaking style $\mathbf{s}_k$ of the $k$-th subject. The generated motion $\mathbf{x}^{1:T}= \left(\mathbf{x}_1, \cdots \mathbf{x}_T \right) \in \mathbb{R}^{T \times V \times 3} $ is a sequence of vertices bias $\mathbf{x}^i \in \mathbb{R}^{V \times 3}$ over a template face mesh comprising $V$ vertices. The audio  $\mathbf{a}^{1:T} = \left( \mathbf{a}_1, \cdots \mathbf{a}_T \right) $ is a speech snippet where $\mathbf{a}_i \in \mathbb{R}^D$ is a audio clip that accounts for the generation of one frame of motion. The speaking style $\mathbf{s}_k \in \mathbb{R}^{K}$ is a one-hot embedding indicating the $k$-th subject from a total of $K$ subjects. We formulate Speech-driven 3D facial animation as a conditional generation problem and propose to use Diffusion model to solve it. Our goal is to generate facial motions $\mathbf{x}^{1:T}$ conditioned on the speech $\mathbf{a}^{1:T}$ and style $\mathbf{s}_{k}$ by sampling from the posterior distribution $p\left(\mathbf{x}^{1:T} \mid \mathbf{a}^{1:T}, \mathbf{s}_k\right)$. For simplicity, we will denote the facial motion and speech as $\mathbf{x}$ and $\mathbf{a}$ in the following, since our model generates the full sequence of motions $\mathbf{x}^{1:T}$ in parallel using the entire speech snippet $\mathbf{a}^{1:T}$ as conditioning context. The distribution $p\left(\mathbf{x}\mid\mathbf{a},\mathbf{s}_k\right)$ accounts for the mapping from given speech snippt and speaking style to all feasible facial motion outputs. To attain this distribution, we refer to Diffusion models and transform a pure Gaussian distribution to it via iterative denoising. Specifically, we establish a Markov chain $p\left(\mathbf{x}_{n-1}\mid\mathbf{x}_{n}, \mathbf{a}, \mathbf{s}_{k}\right)$ to sequentially transform an intermediate distribution $p\left(\mathbf{x}_{n}\right)$ into the next one $p\left(\mathbf{x}_{n-1}\right)$, under the guidance of $\mathbf{a}$ and $\mathbf{s}_k$. Here, $\mathbf{x}_{n} \in \mathbb{R}^{T \times K \times 3}$ represents the facial motion $\mathbf{x}$ combined with Gaussian noise, where the amount of noise is determined by the diffusion step $n \in \{1, \cdots, N\}$.  A larger $n$ indicates more Gaussian noise in $\mathbf{x}_{n}$, with $\mathbf{x}_{N}$ being pure Gaussian noise and $\mathbf{x}_{0}$ being the desired facial motion. The Markov chain successively converts high-noise $\mathbf{x}_{n}$ into lower-noise versions, until reaching the facial motion distribution, through:
    \begin{equation}
        p\left(\mathbf{x}_0 \mid \mathbf{a}, \mathbf{s}_k\right) = p\left(\mathbf{x}_N\right) \prod_{n=1}^T p\left(\mathbf{x}_{n-1} \mid \mathbf{x}_n, \mathbf{a}, \mathbf{s}_k \right), 
    \end{equation}
where $p\left(\mathbf{x}_N\right) = \mathcal{N}(0, \textit{I})$. Since $\mathbf{x} = \mathbf{x}_0$, $p\left(\mathbf{x}_0 \mid \mathbf{a}, \mathbf{s}_k\right)$ is our desired distribution. To infer the less-noised distribution $p\left(\mathbf{x}_{n-1}\right)$ from  $p\left(\mathbf{x}_{n}\right)$ under the guidance of $\mathbf{a}$ and $\mathbf{s}_k$, we refer to a neuron network $\emph{G}$ which is formulated as:
 \begin{align}
     \hat{\mathbf{x}}_0 = \emph{G}(\mathbf{x}_n, \mathbf{a}, \mathbf{s}_k, n),
\label{eq:function}
 \end{align}
$\emph{G}$ acts as a denoiser, which estimates and recovers the facial motion $\hat{\mathbf{x}}_0$ from $\mathbf{x}_n$, conditioned on the audio $\mathbf{a}$, speaking style $\mathbf{s}_k$ and the diffusion step $n$. The estimation $\hat{\mathbf{x}_0}$ is then used to construct the distribution $p\left(\mathbf{x}_{n-1}\right)$ for the next step in the Markov chain, where the exact formulation of $p(\mathbf{x}_{n-1})$ depends on the sampling strategy used~\cite{ho2020denoising,song2020denoising}. We implement the one proposed by~\cite{song2020denoising} since it allows for a relatively short Markov chain, which enables efficient generation.

\begin{figure*}[t]
\centering
\includegraphics[width=1\textwidth]{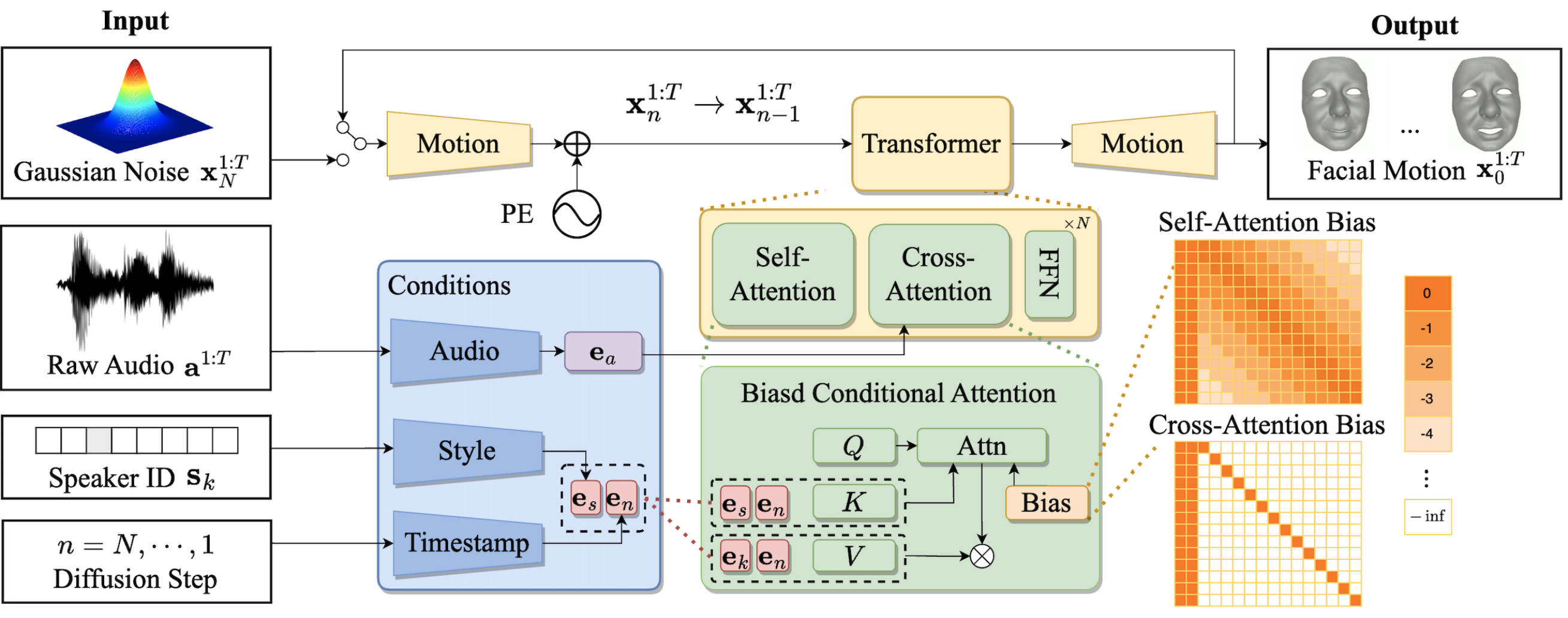} 
\caption{DiffSpeaker synthesizes facial motions $\mathbf{x}^{1:T}$ from speech audio $\mathbf{a}^{1:T}$ and a subject's speaking style $\mathbf{s}_k$, utilizing a Diffusion-based iterative denoising technique. Its core feature is a biased conditional attention mechanism that introduces static biases in self/cross-attention and employs encodings $\mathbf{e}_s$ and $\mathbf{e}_n$ to integrate speaking style and diffusion step information.}
\label{Fig:Method Overview}
\end{figure*}

\begin{figure*}[t]
\centering
\includegraphics[width=1\textwidth]{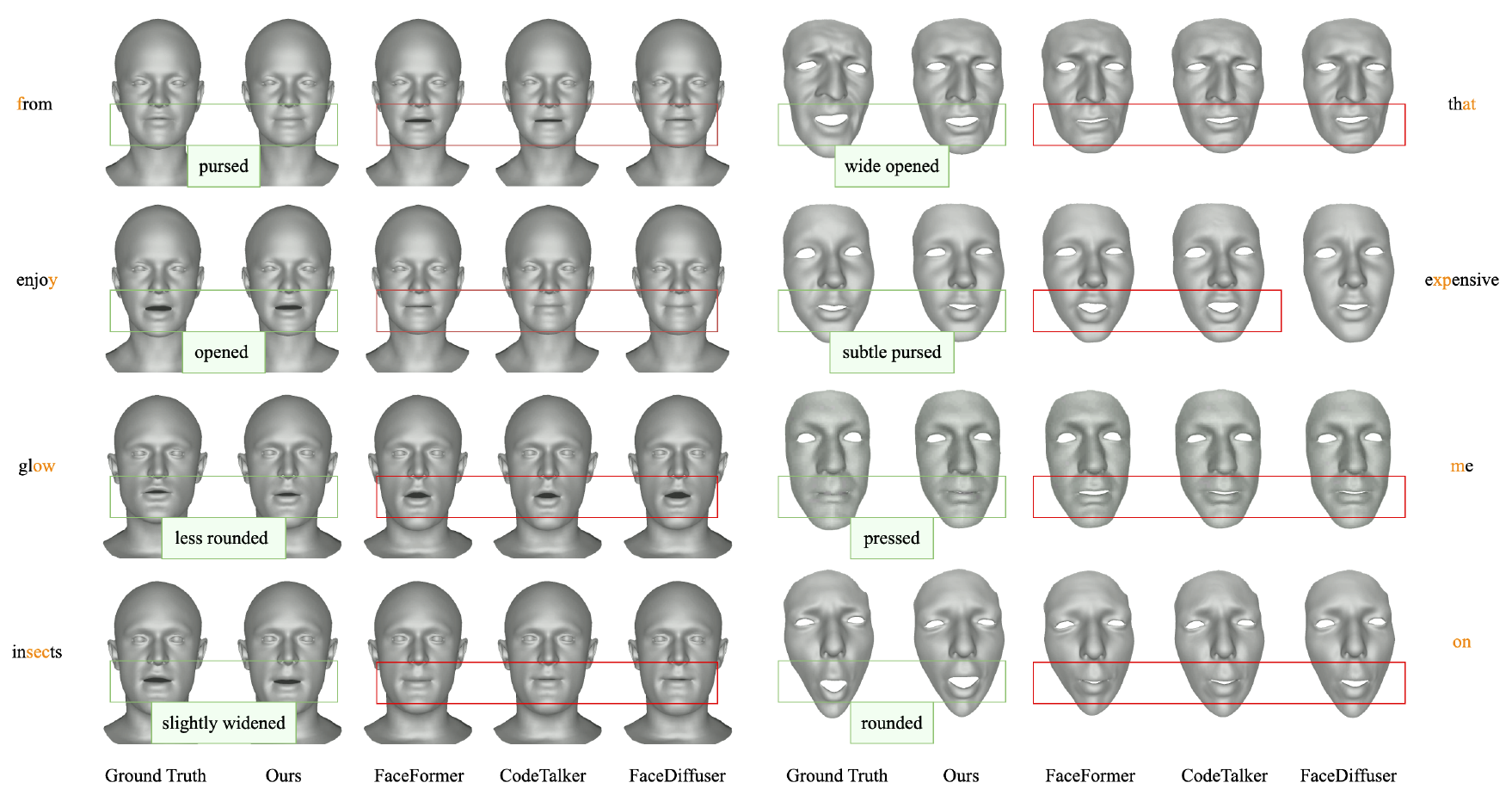} 
\caption{Qualititive Comparision}
\label{Fig:qualititive comparison}
\end{figure*}

\subsection{Diffusion-based Transformer Architecture}

Next, we detail how the conditions for speech $\mathbf{a}$, style $\mathbf{s}_k$, and diffusion step $n$ are incorporated into our Transformer architecture. We use an encoder-decoder architecture for our network $\emph{G}$. As shown in Figure~\ref{Fig:Method Overview}, the conditions are separately encoded by $\emph{E}_a$, $\emph{E}_s$, $\emph{E}_n$ and fed to the Decoder $\emph{D}$ which denoises the input:
\begin{equation}
    \begin{aligned}
        \hat{\mathbf{x}}_0 &= \emph{G}(\mathbf{x}_n, \mathbf{a}, \mathbf{s}_k, n) \\
            &= \emph{D}(\mathbf{x}_n, \emph{E}_a(\mathbf{a}), \emph{E}_s(\mathbf{s}_k), \emph{E}_n(n)),
    \end{aligned}
\end{equation}
where the audio encoder $\mathbf{e}_a = \emph{E}_a(\mathbf{a}^{1:T}) \in \mathbb{R}^{T\times C}$ is a pre-trained audio encoder, style encoder $\mathbf{e}_s = \emph{E}_s(\mathbf{s}_k) \in \mathbb{R}^{1\times C}$ is a linear projection layer, and step encoder $\mathbf{e}_n = \emph{E}_n(n) \in \mathbb{R}^{1 \times C}$ first converts the scalar $n$ into frequency encoding then passes it through a linear layer.  All encoders map their inputs to the same channel dimension $C$. Importantly, our network $\emph{G}$ processes all frame steps $t=\{1,\cdots,T\}$ in parallel, but varies in the diffusion step $n$. 

\subsubsection{Attention with Condition Tokens} 

To address the challenge of processing noise-affected facial motion input with data-hungry attention mechanisms, we incorporate the speaking style condition and the diffusion step into both self-attention and cross-attention layers. Self-attention traditionally enables the facial motion sequence to self-reflect, whereas cross-attention mechanisms integrate the conditional inputs. Given that the facial motion input $\mathbf{x}_n$ is noise-infused, self-attention must be diffusion step-aware, denoted by $n$, to handle the motions accurately. Moreover, since step encoding $\mathbf{e}_n$ and style encoding $\mathbf{e}_s$ are both one-dimensional embeddings, we introduce them concurrently as the condition tokens to the self-attention and cross-attention layers. Adopting the standard Transformer architecture, the outputs from the self-attention layer ($\mathbf{A}_s$) and the cross-attention layer ($\mathbf{A}_c$) are calculated as follows:

\begin{equation}
    \begin{aligned}
        \mathbf{A}_s &= \operatorname{Attention}\left(\mathbf{Q}_s, \left[ \mathbf{e}_s; \mathbf{e}_n; \mathbf{K}_s\right], \left[\mathbf{e}_s; \mathbf{e}_n; \mathbf{V}_s\right]\right), \\
        \mathbf{A}_c &= \operatorname{Attention}\left(\mathbf{Q}_c, \left[\mathbf{e}_s; \mathbf{e}_n; \mathbf{K}_c\right], \left[\mathbf{e}_s; \mathbf{e}_n; \mathbf{V}_c\right]\right),
    \end{aligned}
\end{equation}
$\mathbf{Q}_s$, $\mathbf{K}_s$, $\mathbf{V}_s \in \mathbb{R}^{T \times C}$ for self-attention, and $\mathbf{Q}_c$, $\mathbf{K}_c$, $\mathbf{V}_c \in \mathbb{R}^{T \times C}$ for cross-attention symbolize the query, key, and value features, respectively. The square brackets denote concatenating along the sequence dimension. Prior to their introduction into the attention mechanisms, the style encoding $\mathbf{e}_s$ and the step encoding $\mathbf{e}_n$ are appended to the key, value features. 

\subsubsection{Static Attention Bias}

We introduce fixed biases for both self-attention and cross-attention mechanisms that are specifically crafted to work with condition tokens. To describe this process simply, we refer to the attention operation that incorporates conditional tokens and a static bias as \textit{Biased Conditional Attention}. This mechanism can be applied to both self-attention and cross-attention. Considering an input sequence of length $T$, without loss of generality, assume the number of heads in the attention is 1, our method computes the attention scores for the $i$-th query, $\mathbf{q}_i \in \mathbb{R}^{1 \times C}$, with $i$ ranging from $1$ to $T$, pertinent to either $\mathbf{Q}_s$ or $\mathbf{Q}_c$. Likewise, $\mathbf{K}$ corresponds to either $\mathbf{K}_s$ or $\mathbf{K}_c$. The attention score calculation is as follows:
\begin{equation}
\operatorname{Softmax}\left(\mathbf{q}_i\left[\mathbf{e}_s, \mathbf{e}_n, \mathbf{K}\right]^{T} + \mathbf{b}_i\right),
\end{equation}
where $\mathbf{b}_i$ represents the unique bias term for either self-attention or cross-attention mechanisms.

\textbf{Biased Conditional Cross-Attention}. The cross-attention mechanism facilitates the interaction between facial motion sequences and the associated conditions, which include speech representations $\mathbf{e}_a$, style encoding $\mathbf{e}_s$, and diffusion step encoding $\mathbf{e}_n$. For an attention map of size $T\times(T+2)$, which includes $T$ tokens for facial motion and $2$ tokens for additional conditions, for the $i$-th query, we denote $\mathbf{b}_i(j)$ as the attention bias for the $j$-th value, where $j$ ranges from $1$ to $T+2$. The cross-attention bias is formulated as follows:
\begin{equation}
    \mathbf{b}_i(j)= \begin{cases}
                                0, & j \in \{1, 2, i\}, \\
                                -\infty, & \text{otherwise},
                    \end{cases}
\end{equation}
as depicted in Figure~\ref{Fig:Method Overview}, the design is set up such that the facial motion of a specific frame is limited to engage only with its corresponding speech representation, as well as with $\mathbf{e}_s$ and $\mathbf{e}_n$. This constraint ensures the delivery of audio information that is synchronized in time, while also incorporating information about the diffusion step and speaking style.

\textbf{Biased Conditional Self-Attention}. In the self-attention map of shape $T\times(T+2)$, the bias $\mathbf{b}_i(j)$ is formulated as:
\begin{equation}
    \mathbf{b}^s_i(j)= \begin{cases}
                                0, & 1 \leq j \leq 2, \\
                                \lfloor(i-j) / \mathbf{p}\rfloor, & 2 < j \leq i, \\ 
                                \lfloor(j-i) / \mathbf{p}\rfloor, & i < j \leq T+2 ,
                    \end{cases}
\end{equation}
where $\mathbf{p}$ stands for the constant that is equivalent to the frame rate of the facial motion sequences. As depicted in Figure~\ref{Fig:Method Overview}, this consistent bias is zero for the $2\mathbf{p}$ elements nearest to the diagonal and diminishes as the separation from the diagonal increases. Notably, the bias is anchored at zero for the initial two columns. The interval $\mathbf{p}$ equips the bias with the capability to decode facial motion across a consistent time span, regardless of the frame rate. This bias mitigates the disruptive effects of noise in the facial motion sequences by constraining the self-attention to a focused range. Simultaneously, it preserves the guidance of diffusion step and speaking style.

\textbf{Discussion with FaceFormer}. Our static bias takes cues from FaceFormer~\cite{fan2022faceformer}, which introduced a similar bias for an auto-regressive Transformer that produces sequences incrementally, one element at a time. However, our context differs as we employ a Diffusion-based generative model that creates sequences all at once, synchronously, and these sequences are initially disrupted by Gaussian noise. To adapt to this, we refine the design to enhance the ability of self-attention and cross-attention  to handle full-length noisy sequences. The effectiveness of this modification is demonstrated in the experimental section.

\subsection{Training Objective.} We train the entire network in Figure~\ref{Fig:Method Overview}, including the pre-trained audio encoder, with the objective of recovering the original signal from any diffusion step $n \in \{1, \cdots, N \} $. For any $\mathbf{a}, \mathbf{s}_k, \mathbf{x}_0$ sampled from the dataset, we have the following loss to supervise the recovered $\hat{\mathbf{x}_0} = \emph{G}(\mathbf{x}_n, \mathbf{a}, \mathbf{s}_k, n)$:
$$
\mathcal{L}_{r e c}=\mathbb{E}_n\left[\frac{1}{T} \sum_{t=1}^T\left\|\mathbf{x}_0^t-\hat{\mathbf{x}}_0^t\right\|^2\right],
$$
where $\mathbf{x}^{t}_{0}$ and $\hat{\mathbf{x}}^{t}_{0}$ are the $t$-th frames of the ground truth $\mathbf{x}_0$ and prediction $\hat{\mathbf{x}}_0$. We also use a velocity loss to address jitter and encourage smooth motions over time:
$$
\mathcal{L}_{v e l}=\mathbb{E}_n\left[\frac{1}{T} \sum_{t=1}^T\left\|\left(\mathbf{x}_0^{t-1}-\mathbf{x}_0^t\right)-\left(\hat{\mathbf{x}}_0^{t-1}-\hat{\mathbf{x}}_0^t\right)\right\|^2\right],
$$

The total loss is the sum of the two losses:

\begin{equation}
\mathcal{L}= \lambda_1 \mathcal{L}_{rec}+ \lambda_2 \mathcal{L}_{vel},
\end{equation}

where $\lambda_1 = \lambda_2 = 1$. This objective aligns with our baselines~\cite{cudeiro2019capture,fan2022faceformer}.

\section{Experiments}
\subsection{Datasets and Implementations}

We conduct training and testing with two open-source 3D facial datasets, BIWI~\cite{fanelli20103} and VOCASET~\cite{cudeiro2019capture}. Both datasets have 4D face scans along with audio recordings of short sentences.

\textbf{BIWI dataset.} The BIWI dataset contains recordings of 40 unique sentences spoken by 14 subjects - 8 females and 6 males. Each subject speaks each sentence twice - once emotionally and once neutrally. The average duration of each sentence recording is 4.67 seconds. The recordings were captured at 25 frames per second with 23370 vertices per 3D face scan frame. Following precedents set by previous work~\cite{fan2022faceformer,xing2023codetalker}, we only utilized the emotional split, and divided it into three subsets, a training set (BIWI-Train) with 192 sentence recordings, a validation set (BIWI-Val) with 24 sentences, a test set A (BIWI-Test-A) with 24 sentences spoken by 6 subjects seen during training, and a test set B (BIWI-Test-B) with 32 sentences spoken by 8 novel subjects not seen during training. We use BIWI-Test-A for both quantitative and qualitative evaluation. BIWI-Test-B is used primarily for qualitative testing.

\textbf{VOCASET dataset.} The VOCASET dataset contains 480 paired audio and 3D facial motion sequences captured from 12 subjects. The facial motion sequences are recorded at 60 frames per second and are approximately 4 seconds in length. Unlike BIWI, the 3D face meshes in VOCASET are registered to the FLAME~\cite{li2017learning} topology, which has 5023 vertices per mesh. We adopted the VOCASET into three subsets, a training set (VOCA-Train), a validation set (VOCA-Val), and a test set (VOCA-Test) as in ~\cite{fan2022faceformer,xing2023codetalker} for fair comparison.

\begin{table}{
\setlength{\tabcolsep}{0.02cm}
\begin{tabular}{@{}lccc@{}}
\toprule
\multirow{2}{*}{Methods} & \multicolumn{2}{c}{BIWI}                                                                                                                                                                                    & \multicolumn{1}{c}{VOCASET}                                                                                              \\ \cmidrule(l){2-4} 
                         & \begin{tabular}[c]{@{}c@{}}LVE $\downarrow$\\ \multicolumn{1}{l}{($\times 10^{-4}$mm)} \end{tabular} 
                         & \begin{tabular}[c]{@{}c@{}}FDD $\downarrow$\\ \multicolumn{1}{l}{($\times 10^{-5}$mm)} \end{tabular} 
                         & \multicolumn{1}{l}{\begin{tabular}[c]{@{}c@{}}LVE $\downarrow$\\ \multicolumn{1}{l}{($\times 10^{-5}$mm)} \end{tabular}} \\ \midrule
VOCA                     & 6.5563                                                                                               & 8.1816                                                                                               & 4.9245                                                                                                                   \\
MeshTalk                 & 5.9181                                                                                               & 5.1025                                                                                               & 4.5441                                                                                                                   \\
FaceFormer$^{\left[1\right]}$               & 5.3077                                                                                               & 4.6408                                                                                               & 4.1090                                                                                                                   \\
CodeTalker$^{\left[1\right]}$               & 4.7914                                                                                                                                                                            & 4.1170                                                                                               & 3.9445                                                                                                                                                                                         \\ 


FaceDiffuser$^{\left[2\right]}$               & 4.7823                                                                                              & 3.9225                                                                                               & 4.1235                                                                                                                 \\  \midrule

DiffSpeaker$^{\left[1\right]}$ (Ours)        & 4.5556                                                                                               & \textbf{3.6823}                                                                                              &  3.2213                                                                                                               
                    \\ 

DiffSpeaker$^{\left[2\right]}$ (Ours)        & \textbf{4.2829}                                                                                             & 3.8535                                                                                             & \textbf{3.1478}                                                                                                                
                    \\ 
                    \bottomrule
\end{tabular}
}
\caption{Quantitative comparison of DiffSpeaker with other state-of-the-art methods. $[1],[2]$ denote pre-trained audio encoders, as elaborated in the implementation section.}
\label{tab:quantitative_comparsion}
\end{table}

\subsection{Implementation Detail}

\textbf{Network Architecture}. During our evaluations on the VOCASET dataset \cite{cudeiro2019capture}, we implemented a Transformer model with a configuration that includes 512 dimensions for hidden states, 1024 dimensions for feedforward networks, alongside 4 heads for multi-head attention, and a single Transformer block. The self/cross-attention in the Transformer utilize residual connections, meaning that the output of each attention operation is summed with its input. For experiments with the BIWI dataset \cite{fanelli20103}, which presents more intricate data, model is scaled up by enhancing the hidden state dimensions to 1024 and the feedforward network dimensions to 2048, other configurations keep consistent with those outlined in VOCASET. This design strategy is consistent with established models from related literature \cite{fan2022faceformer}, which have demonstrated effective Transformer models tailored to these particular datasets.

\textbf{Training}. Our model was developed with the PyTorch framework and Nvidia V100 GPUs. Specifically, when training on the VOCASET dataset, we used a batch size of 32 and ran the training on a single V100 GPU, which took approximately 24 hours to complete. In contrast, for the BIWI dataset, the training was distributed across 8 V100 GPUs, with each GPU processing a batch size of 32, culminating in a total training time of 12 hours. Throughout all training sessions, we employed the AdamW optimization algorithm, setting the learning rate to 0.0001.

\textbf{Baselines.} We compare our work against state-of-the-art methods, including VOCA~\cite{cudeiro2019capture}, MeshTalk~\cite{richard2021meshtalk}, FaceFormer~\cite{fan2022faceformer}, CodeTalker~\cite{xing2023codetalker} and FaceDiffuser~\cite{stan2023facediffuser} on BIWI and VOCASET datasets. While additional dataset and cross-modal support are known to enhance performance, as seen in other tasks, our investigation is limited to network-focused studies without such support. For fair comparison, we use pre-trained audio encoders follow existing methods, referenced as [1]~\cite{baevski2020wav2vec} and [2]~\cite{hsu2021hubert}. For tests involving audio from previously unheard speakers, we traverse all speaking styles learned during training. In probabilistic mapping experiments, we compute the average results from 10 random seeds.


\subsection{Quantitative Evaluation}

\begin{table*}[ht]
\begin{center}
\centering
\begin{tabular}{@{}llccc@{}}
\toprule
\multicolumn{2}{l}{\multirow{2}{*}{}} & \multicolumn{2}{c}{BIWI} & \multicolumn{1}{c}{VOCASET} \\ \cmidrule(l){3-5} 
\multicolumn{2}{l}{}                  & \begin{tabular}[c]{@{}c@{}}LVE $\downarrow$\\ \multicolumn{1}{l}{($\times 10^{-4}$mm)} \end{tabular}            & \begin{tabular}[c]{@{}c@{}}FDD $\downarrow$\\ \multicolumn{1}{l}{($\times 10^{-5}$mm)} \end{tabular}            & \multicolumn{1}{l}{\begin{tabular}[c]{@{}c@{}}LVE $\downarrow$\\ \multicolumn{1}{l}{($\times 10^{-5}$mm)} \end{tabular}}     \\ \midrule
\multicolumn{2}{l}{Ours}              & $\textbf{4.5556} \pm \textbf{9.4441} \times 10^{-7} $    & $3.6823 \pm 4.8427 \times 10^{-6}$     & $\textbf{3.2213} \pm \textbf{2.2093} \times 10^{-7}$  \\ \cmidrule(l){1-5} 
    \multicolumn{1}{c}{Guidence =} & 0.0           & $4.8111 \pm 2.6001 \times 10^{-6}$      & $2.8073 \pm 5.7399 \times 10^{-6}$       & $3.2416 \pm 2.4803 \times 10^{-7}$     \\
                      & 0.5         & $5.1918  \pm 4.0156 \times 10^{-6}$       & $2.8299 \pm 9.2897 \times 10^{-6}$     & $3.5202 \pm 3.9055 \times 10^{-7}$    \\
                      & 1.0           & $6.2146 \pm 4.9915 \times 10^{-4}$      & $\textbf{1.6247} \pm 2.3166 \times 10^{-4}$    & $4.7061 \pm  8.5118 \times 10 ^ {-7}$    \\ \cmidrule(l){1-5} 
\multicolumn{2}{l}{w/o Cross-Bias}    & $11.341 \pm 5.5291 \times 10^{-4}$      & $5.2064 \pm 9.1984 \times 10^{-3}$       & $5.1911 \pm 2.2548 \times 10^{-7}$    \\
\multicolumn{2}{l}{w/o Self-Bias}     & $4.5963 \pm 1.5417 \times 10^{-6}$      & $4.0871 \pm \textbf{4.7425} \times 10^{-6}$      & $3.2880 \pm 2.7058 \times 10^{-7}$    \\ 
\multicolumn{2}{l}{w/o Cond-Self-Attn} & $4.8732 \pm 3.7476 \times 10^{-6}$      & $4.8537 \pm 1.9346 \times 10^{-5}$     & $3.7453 \pm 4.6903 \times 10^{-7}$         \\ 
\multicolumn{2}{l}{FaceFormer-Bias}  & $4.6963 \pm 1.4362 \times 10^{-6}$      & $4.4572 \pm 8.1231 \times 10^{-6}$ & $3.5201 \pm 4.7034 \times 10^{-7}$         \\ 
\multicolumn{2}{l}{Fully Self-Attn}  & $ 9.9793 \pm 8.7841 \times 10^{-5}$      & $4.3232 \pm 8.6785 \times 10^{-5}$ & $4.6816 \pm 5.4681 \times 10^{-7}$         \\ 
\bottomrule
\end{tabular}
\end{center}
\caption{Ablation study of DiffSpeaker. Both mean values and $95\%$ confidential intervals are included.}
\label{tab:ablation study}
\end{table*}

Our evaluation assesses lip sync accuracy and facial expression naturalness. Lip synchronization is gauged using the lip vertex error (LVE) metric from \cite{richard2021meshtalk}, while the naturalness is measured by the facial dynamics deviation (FDD) from \cite{fan2022faceformer}. We apply these metrics to the BIWI dataset and only LVE for the VOCASET, based on its limited facial expression variation as indicated by \cite{xing2023codetalker}. In speech-driven 3D facial animation, lip sync reflects how well the animation corresponds with the audio, and facial expression variation captures the expressive naturalness loosely related to speeches. Table~\ref{tab:quantitative_comparsion} presents that our method achieves strong audio-visual alignment with diverse facial expressions. When contrasted with FaceDiffuser, the improvement indicates that integrating  Transformer structure with Diffusion-based generation improves lip alignment without compromising the stochastic component of facial movements; comparing to other methods, it underscores the suitability of conditional probabilistic approaches, given the intrinsic  one-to-many nature of this task.

\begin{figure}[t]
\centering
\includegraphics[width=0.7\linewidth]{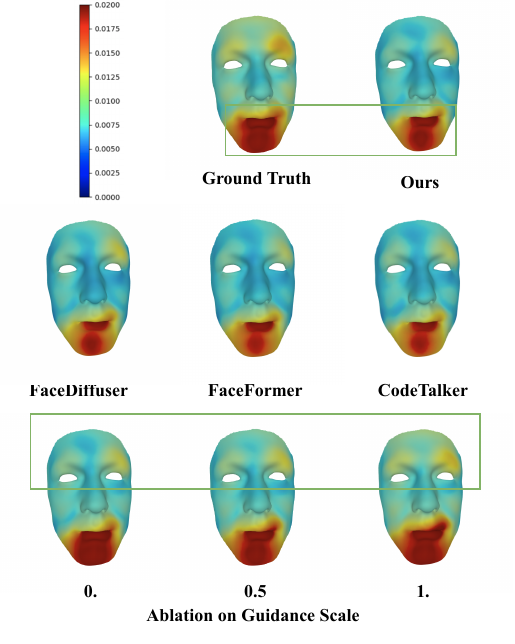} 
\caption{Standard deviation of facial motion. }
\label{Fig:visualization of devaiation}
\end{figure}

\subsection{Qualitative Evaluation}
We visually compare our method with FaceFormer, CodeTalker, and FaceDiffuser. Figure~\ref{Fig:qualititive comparison} shows the animation results of different methods. Compared to other methods, our model generates more accurate lip movements than other methods. We observe improved accuracy in anticipating pronunciations, like pursed lips before the "from" and mouth opening before "on". The lip movements are also more accurate after pronunciations, like the mouth not fully closing after "that" and lips pouting after "glow". Moreover, our model shows improved synchronization when transitioning between syllables, like mouth opening between "in" and "sects" or closing for "x" in "expensive". We recommend watching the supplementary video for more detail. The naturalness of generated facial animations can be inferred to some extent by visualizing the standard deviation of facial movements across the test dataset. As illustrated in Figure~\ref{Fig:visualization of devaiation}, our method more closely replicates the range of facial motion variations when contrasted with alternative approaches.

\subsection{Inference Latency}

\begin{figure}[t]
\centering
\includegraphics[width=1\linewidth]{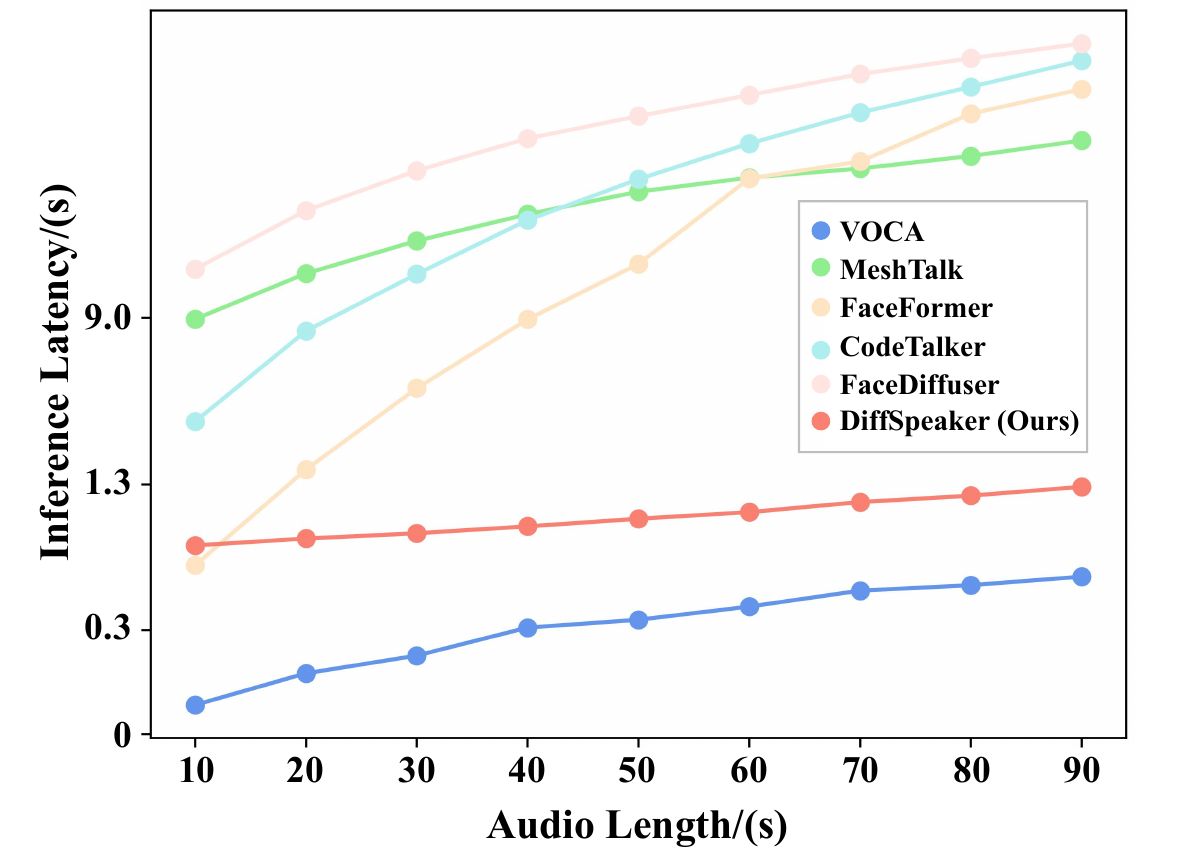} 
\caption{Inference latency for 10-90 second audio clips.}
\label{Fig:latency}
\end{figure}

Despite the common perception of Diffusion-based methods as slow, our approach for speech-driven 3D facial animation demonstrates faster inference speeds than most current alternatives. Latency was measured for audio of varying lengths on a 3090 GPU, with results in Figure~\ref{Fig:latency} showing our method surpassing all others except VOCA for audio over 10 seconds. This speed benefit becomes more pronounced with longer audio clips due to our use of fixed denosing iterations regardless of audio length. Unlike VOCA, which processes all audio windows in parallel, other methods depend on sequentially attending to preceding segments for subsequent predictions, leading to increased iterations for extended audios.
\subsection{Ablation Study}


\textbf{Impact of the Unconditional Guidance}. Conditional diffusion-based models often employ a hybrid approach that mixes conditional and unconditional denoising in the reverse generation process, as per \cite{ho2022classifier}. This method allows for samples that not only adhere to the set conditions but also possess greater variability. The model is trained to occasionally denoise without the condition, with a certain probability set during training. At inference, the guidance scale parameter $w$ adjusts the mix of conditional (weighted by $1+w$) and unconditional (weighted by $-w$) denoising steps. Following the protocol in \cite{chen2023executing}, our model is trained for unconditional denoising with a $10\%$ chance by nullifying the audio input. Table~\ref{tab:ablation study} shows that increasing the guidance scale $w$ from $0$ to $1$ leads to increased Lip Video Error (LVE), indicating worse lip-sync with speech, even though facial expression is more natural (as per Facial Detail Distance, FDD). Figure~\ref{Fig:visualization of devaiation} illustrates the variation of facial movements across the test dataset. With the application of unconditional guidance, there is a notable enhancement in the variation of facial expressions, as measured by FDD, which come closer to matching the variability of real human motions. This increase in variance is attributed to the inclusion of unconditional elements that contribute to aspects of facial movement not directly linked to the audio signal. While this inclusion yields more pronounced expressions, particularly in the upper face, it simultaneously disrupts the synchronization between audio and lip movements. Given that our task prioritizes the accuracy of audio-lip alignment, which we believe is crucial for maintaining the integrity of audio-visual congruence, we opted not to implement this guidance approach. Consequently, our results remain stable across various random seeds, as detailed by the confidence intervals in Table~\ref{tab:ablation study}.


\textbf{Impact of the Transformer Architecture}. We evaluate the specific design of conditional biased self/cross-attention within a diffusion-based Transformer. The cross-attention bias is key for achieving precise temporal alignment between the audio inputs and facial animation outputs. Removing this bias from the model, indicated as \textbf{w/o Cross-Bias} in Table~\ref{tab:ablation study} results in substantially worse performance on both the Lip vertex error (LVE) and facial dynamics deviation (FDD). They suggest the decrease in the synchronization quality between lip movements and the spoken words, as well as in the facial expression. Besides, the \textbf{w/o Self-Bias} row in Table~\ref{tab:ablation study} demonstrates that the inclusion of self-attention bias enhances model performance, suggesting the advantage of having facial motions give greater consideration to adjacent frames. This observation aligns with FaceFormer~\cite{fan2022faceformer} which implemented similar self/cross-attention biases within an auto-regressive Transformer framework. To evaluate the significance of our tailored biases within the Diffusion-based Transformer, we replaced the self/cross attention biases with those from FaceFormer , as indicated in the \textbf{FaceFormer-Bias} row of Table~\ref{tab:ablation study}. This substitution results in biases that equally affect the speaking style and diffusion step conditions, alongside the facial motion sequences. This uniform biasing does not offer effective condition guidance to extended sequences, leading to sub-optimal performance. On the other hand, excluding the conditions of speaking style and diffusion step from the self-attention mechanism, or adopting a decoder-only Transformer architecture where all attention layers concurrently process both input and output, leads to inferior results. These setups are denoted as \textbf{w/o Cond-Self-Attn} and \textbf{Fully Self-Attn} in Table~\ref{tab:ablation study}. While such configurations have been successfully applied in other generative tasks, they typically depend on the availability of large datasets. This is because the self-attention layers need to either handle sequences with significant noise, or process the input and output as a single, integrated sequence. Due to the scarcity of paired speech-4D data, our architecture effectively mitigates the data-hungry nature of Transformers and leverages their potential for high performance.

\section{Conclusion}


In this work, we explored effectively combining Transformer architecture with Diffusion-based framework for the purpose of speech-driven 3D facial animation. The core of our contribution is the introduction of the biased conditional self/cross-attention mechanism, which tackles the difficulties of training a diffusion-based Transformer with constrained and short-spanned audio-4D data. We also investigated the balance between achieving accurate lip sync and generating facial expressions with less speech correlation. The model we developed surpasses current methods, excelling in both the quality of animation and the speed of generation.




\bibliographystyle{named}
\bibliography{ijcai24}

\clearpage

\end{document}